\documentclass{article}
\usepackage{spconf,amsmath,amssymb,amsfonts,graphicx,booktabs}
\usepackage{stmaryrd}
\usepackage{xcolor}
\usepackage{flushend}

\usepackage{enumitem}
\setlist[itemize]{%
labelsep=4pt,%
labelindent=0.2\parindent,%
itemindent=0pt,%
leftmargin=*,%
itemsep=-1pt,
}

{}

\title{Learning filterbanks from raw speech for phone recognition}
\name{
Neil Zeghidour$^{1,2}$, Nicolas Usunier$^1$, Iasonas Kokkinos$^1$, Thomas Schatz$^2$,
\vspace{-0.4cm}
}
\address{
\emph{Gabriel Synnaeve}$^1$, \emph{Emmanuel Dupoux}$^2$\\
~\\
$^1$ Facebook A.I. Research, Paris, France, New York, USA\\
$^2$ CoML, ENS/CNRS/EHESS/INRIA/PSL Research University, Paris, France
}

\begin{document}
%\ninept
%
\maketitle
\begin{abstract}
We train a bank of complex filters that operates on the raw waveform and is fed into a convolutional neural network for end-to-end phone recognition. These time-domain filterbanks (TD-filterbanks) are initialized as an approximation of mel-filterbanks% 
%(MFSC, for mel-frequency spectral coefficients)
, and then fine-tuned jointly with the remaining convolutional architecture. We perform phone recognition experiments on TIMIT and show that for several architectures, models trained on TD-filterbanks consistently outperform their counterparts trained on comparable mel-filterbanks%
%MFSC
. We get our best performance by learning all front-end steps, from pre-emphasis up to averaging. Finally, we observe that the filters at convergence have an asymmetric impulse response, and that some of them remain almost analytic.
\end{abstract}
\vspace{-6pt}
\vspace{-0.14cm}
\section{Introduction}
\label{sec:intro}
\vspace{-0.2cm}

Speech features such as gammatones or mel-filterbanks (MFSC, for mel-frequency spectral coefficients) were
designed to match the human perceptual system
\cite{davis1980comparison,patterson1987efficient}, and contain
invaluable priors for speech recognition tasks. However, even if a
consensus has been reached on the proper setting of the
hyperparameters of these filterbanks along the years, there is no
reason to believe that they are optimal representations of
the input signal for all recognition tasks. In the same way deep
architectures changed the landscape of computer vision by directly
learning from raw pixels \cite{lecun1998gradient,alexnet}, we believe
that future end-to-end speech recognition system will learn directly
from the waveform.

There have been several attempts at learning directly from the raw
waveform for speech recognition
\cite{palaz1,hoshen2015speech,sainath2015learning,tjandra2017attention}. \cite{hoshen2015speech,sainath2015learning} propose an architecture composed of a convolutional layer followed by max-pooling and a
nonlinearity, so that gammatone filterbanks correspond to a particular
configuration of the network. \cite{tjandra2017attention} explore an alternative architecture, with the intention to represent MFSC rather than gammatones. They propose a $4$-layer convolutional architecture followed by two networks-in-networks \cite{nin}, pretrained to reproduce MFSC.
\begin{figure}

\begin{minipage}[b]{1\linewidth}
  \centering
  \centerline{\includegraphics[width=9cm]{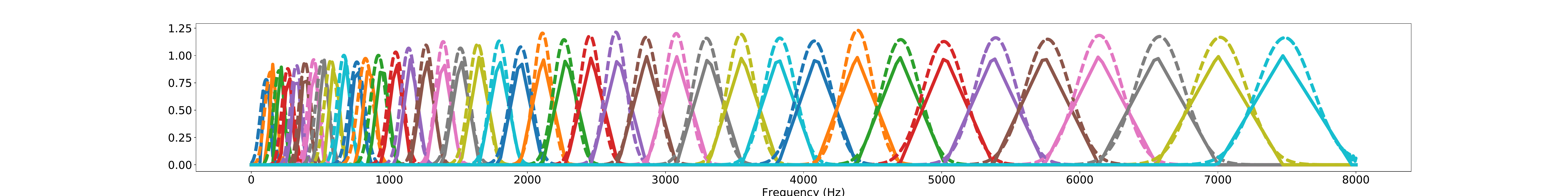}}

  \centerline{(a) MFSC and Gabor filter approximation}%\medskip

\end{minipage}
\begin{minipage}[b]{\linewidth}
  \centering
  \centerline{\includegraphics[trim={0 10cm 0 10.5cm},clip,width=9cm]{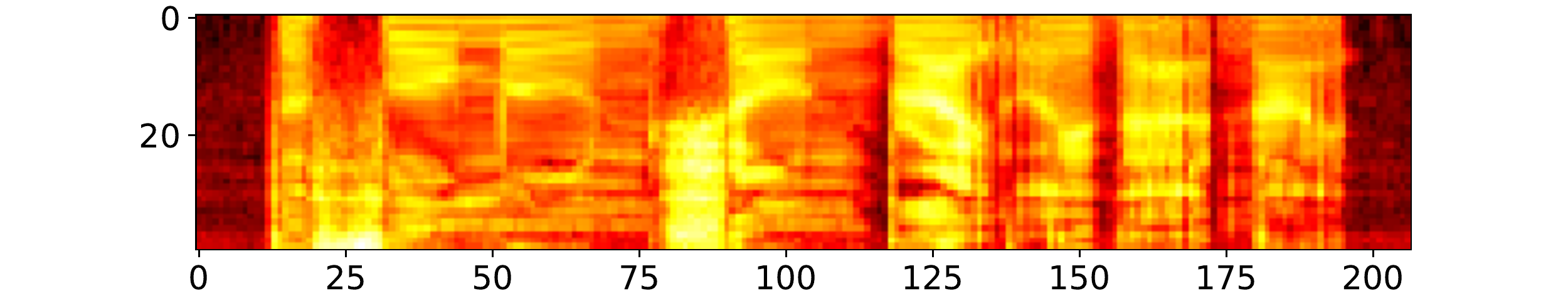}}
  \centerline{(b) MFSC}%\medskip
\end{minipage}
\\
\begin{minipage}[b]{\linewidth}
  \centering
  \centerline{\includegraphics[trim={0 10cm 0 10.5cm},clip,width=9cm]{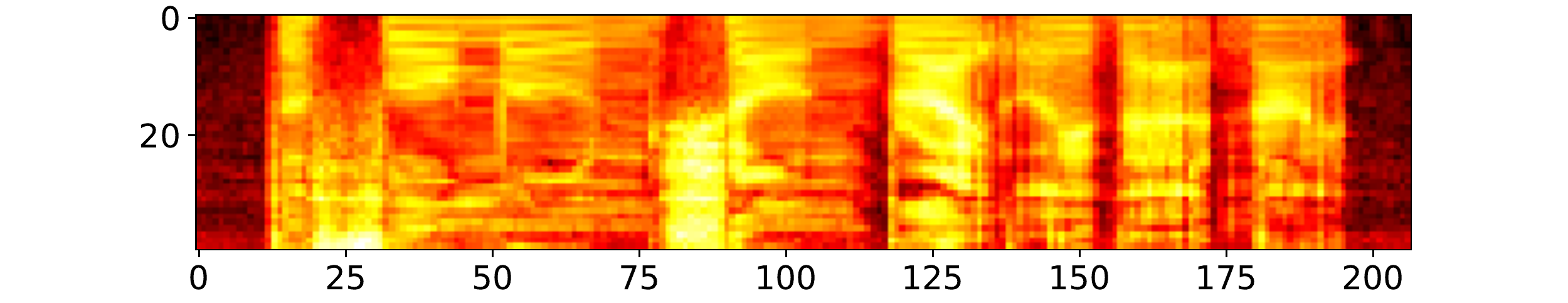}}
  \centerline{(c) TD-filterbanks with the Gabor filters}%\medskip
\end{minipage}
\vspace{-0.42cm}
\caption{Frequency response of filters, and output of MFSC and their time-domain approximation on a sentence of TIMIT.}
\vspace{-10pt}
\label{fig:fbanks}
\end{figure}

We also focus on MFSC
because they are the front-end of state-of-the-art phone
\cite{laszlo_best_timit} and speech \cite{msr_human_parity}
recognition systems.
Our work builds on \cite{dss}, who introduce a time-domain approximation of MFSC using the first-order coefficients of a
scattering transform. This leads us to study an architecture using a
convolutional layer with complex-valued weights, followed by a modulus
operator and a low-pass filter.
In contrast to \cite{tjandra2017attention}, we propose a lightweight architecture that serves as a plug-in, learnable replacement to MFSC in deep neural networks. Moreover,
we avoid pretraining by initializing the complex convolution weights with Gabor
wavelets whose center frequency and bandwidth match those of MFSC.

We perform phone recognition experiments on TIMIT and show that given
competitive end-to-end models trained with MFSC as inputs, training
the same architectures by replacing the MFSC with the
learnable architecture leads to performances that are better than when using MFSC.
Moreover, our best model is
obtained by learning everything except for the non-linearities, including a
pre-emphasis layer.

\section{Time-domain MFSC}
\vspace{-0.4cm}
\newcommand{\hanw}{s}
We present the standard MFSC and their practical
implementation. We then describe a learnable replacement of MFSC that uses only convolution
operations in time domain, and how to set the weights to reproduce
MFSC.

\vspace{-0.4cm}
\subsection{MFSC computation}
\vspace{-0.4cm}
Given an input signal $x$, MFSC are computed by first
taking the short-time Fourier transform (STFT) of $x$ followed by taking
averages in the frequency domain according to triangular filters with
centered frequency and bandwidth that increase linearly in
log-scale. More formally, let $\phi$ be a Hanning window of width
$\hanw$ and $(\psi_n)_{n=1..N}$ be $N$ filters whose squared frequency response
are triangles centered on $(\eta_n)_{n=1..N}$ with full width at half
maximum (FWHM) $(w_n)_{n=1..N}$. Denoting by $x_t: u \mapsto x(u)\phi(t-u)$ the
windowed signal at time step $t$, and $\hat{f}$ the Fourier transform of function $f$, the filterbank is the set of $N$
functions $(t\mapsto M x(t, n))_{n=1..N}$:
\begin{equation}
  Mx(t,n) = \frac{1}{2\pi} \int |\hat{x}_t(\omega)|^2|\hat{\psi}_n(\omega)|^2d\omega\,,
\end{equation}

\vspace{-0.4cm}
\subsection{Approximating MFSC with convolutions in time}
\vspace{-0.4cm}
As in \cite{dss}, we approximate MFSC
in the time domain using:
\begin{equation}
\label{eq:nfsc}
    Mx(t,n) \approx |x*\varphi_{n}|^2*|\phi|^2(t) .
\end{equation}
where $\varphi_n$ is a wavelet that approximates the $n$-th triangular filter in frequency, i.e.  $|\hat{\varphi}_n|^2 \approx
|\hat{\psi}_n|^2$, while $\phi(t)$ is the Hanning window also used for the MFSC. The approximation is valid when the time support of $\varphi_n$ is smaller than that of $\phi$.

This approximation of MFSC is also known as a first-order scattering transform. This is the foundation of the deep scattering spectrum \cite{dss}, which cascades scattering transforms to retrieve information that is lost in the MFSC. Deep scattering spectra have been used as inputs to neural networks trained for phone recognition \cite{peddinti2014deep} or classification \cite{zeghidour2016deep}, which showed better performances than comparable models trained on MFSC. In this work, we do not use the deep scattering spectrum. First-order scattering coefficients provide us with both a design for the first layers of the network architecture to operate on the waveform, and an initialization that approximates the MFSC computation.

Given the MFSC center
frequencies $(\eta_n)_{n=1..N}$ and FWHM $(w_n)_{n=1..N}$, we use \eqref{eq:nfsc} to approximate MFSC with Gabor wavelets:
\begin{equation}
    \varphi_n(t) \propto e^{-2\pi i \eta_n t} \frac{1}{\sqrt{2\pi}\sigma_n} e^{-\frac{t^2}{2\sigma_n^2}} .
\end{equation}
where $\eta_n$ is the desired center frequency, and the width parameter
$\sigma_n$ of the Gabor wavelet is set to match the desired FWHM
$w_n$. Since for a frequency $\xi$ we have $\hat{\varphi}_n(\xi) \propto
\sqrt{\sigma_n}e^{-\frac{1}{2}\sigma_n^2(\xi-\eta_n)^2}$, the FWHM is
$2\sqrt{2\log{2}}\sigma_n^{-1}$ and we take $\sigma_n =
\frac{2\sqrt{2\log{2}}}{w_n}$. Each $\varphi_n$ is then normalized to
have the same energy as $\psi_n$. Figure~\ref{fig:fbanks} (a) shows in frequency-domain the triangular averaging operators of usual
MFSC and the corresponding Gabor wavelets. Figures~\ref{fig:fbanks}
(b) and (c) compare the $40$-dimensional spectrograms of the MFSC
and the Gabor wavelet approximation on a random sentence of the TIMIT
corpus after mean-variance normalization, showing that the
spectrograms are
similar.

\noindent{\bf MFSC specification.}
The standard setting in speech recognition is to start from the
waveform sampled at $16$kHz and represented as $16$-bit signed
integers. The STFT is computed with $512$ frequency bins using Hanning
windows of width $25$ms, and decimation is applied by taking the STFT
every $10$ms. There are $N=40$ filters, with center frequencies
$(\eta_n)_{n=1..N}$ that span the range $64Hz-8000Hz$ by being equally spaced on a mel-scale. The final
features are the $\log(\max(Mx(t,n), 1))$. In practice, the STFT is
applied to the raw signal after a pre-emphasis with parameter
$0.97$, and coefficients have mean-variance normalization per utterance.

\renewcommand{\arraystretch}{0.8}
\begin{table}
\centering
{\small
\begin{tabular}{lcccc}
\toprule

Layer type & Input size & Output size & Width & Stride\\
\midrule
Conv. & 1 & 80 & 400 & 1 \\
L2-Pooling & 80 & 40 & - & - \\
Square & - & - & - & - \\
Grouped conv. & 40 & 40 & 400 & 160 \\
Absolute value & - & - & - & - \\
Add 1, Log & - & - & - & - \\
\bottomrule
\end{tabular}}
\caption{Details of the layers for the TD-filterbanks.}
\label{tab:nfsc}
\vspace{-2ex}
\end{table}

\begin{table}
\centering
\begin{tabular}{lcc}
\toprule
Learning mode & Dev PER & Test PER \\
\midrule
MFSC & 17.8 & 20.6\\
\cmidrule(lr){1-1}%\midrule
Fixed & 18.3 & 21.8 \\
Learn-all & 17.4 & 20.6 \\
Learn-filterbank & 17.3 & 20.3 \\
Randinit & 29.2 & 31.7 \\
\bottomrule
\end{tabular}
\caption{PER of the CNN-5L-ReLU-do0.7 model trained on MFSC and different learning setups of TD-filterbanks.}
%\end{small}
\label{tab:layer-modes}
\vspace{-8pt}
\end{table}

\begin{table*}
\centering
\begin{tabular}[width=2\coumnwidth]{llcc}
\toprule
Model & Input & Dev PER & Test PER \\
\midrule
Hybrid HMM/Hierarchical CNN + Maxout + Dropout \cite{laszlo_best_timit} &  MFSC + energy + $\Delta$ + $\Delta\Delta$ & 13.3 & 16.5 \\
\midrule
CNN + CRF on raw speech \cite{palaz_baseline} & wav & - & 29.2 \\
Wavenet \cite{wavenet} & wav & - & 18.8 \\
CNN-Conv2D-10L-Maxout \cite{zhang_conv2d} & MFSC & 16.7 & 18.2 \\
Attention model +  Conv. Features + Smooth Focus \cite{attention_based} & MFSC + energy + $\Delta$ + $\Delta\Delta$ & 15.8 & 17.6 \\
LSTM + Segmental CRF \cite{segmental_rnn} & MFSC + $\Delta$ + $\Delta\Delta$ & - & 18.9 \\
LSTM + Segmental CRF \cite{segmental_rnn} & MFCC + LDA + MLLT + MLLR & - & 17.3 \\
\midrule
CNN-5L-ReLU-do0.5 & MFSC & 18.4 & 20.8 \\
CNN-5L-ReLU-do0.5 + TD-filterbanks & wav & 18.2 & 20.4 \\
\cmidrule(lr){1-1}
CNN-5L-ReLU-do0.7 & MFSC & 17.8 & 20.6 \\
CNN-5L-ReLU-do0.7 + TD-filterbanks & wav & 17.3 & 20.3 \\
\cmidrule(lr){1-1}
CNN-8L-PReLU-do0.7 & MFSC & 16.2 & 18.1 \\
CNN-8L-PReLU-do0.7 + TD-filterbanks & wav & 15.6 & 18.1 \\
CNN-8L-PReLU-do0.7 + TD-filterbanks-Learn-all-pre-emp  & wav & 15.6 & 18.0 \\
\bottomrule
\end{tabular}
\caption{PER (Phone Error Rate) on TIMIT, in percentages. All models but \cite{laszlo_best_timit} are trained in an end-to-end fashion.}
\label{tab:results}
%\end{small}
\end{table*}

\noindent{\bf Learnable architecture specification.} The time-domain convolutional
architecture is summarized in Table~\ref{tab:nfsc}. With a waveform
sampled at $16$kHz, a Hanning window is a convolution operator with a
span of $W=400$ samples ($25$ms). Since the energy of the Gabor wavelets
approximating standard MFSC has a time spread smaller than the
Hanning window, the complex wavelet+modulus operations
$|x*\varphi_{n}|^2$ are implemented as a convolutional layer taking
the raw wav as input, with a width $W=400$ and $2N=80$ filters ($40$
filters for the real and imaginary parts respectively). This layer is on the top
row of Table~\ref{tab:nfsc}. The modulus operator is implemented with
``feature L2 pooling'', a layer taking an input $z$ of size $2N$ and
outputs $z'$ of size $N$ such that $z'_k = \sqrt{z_{2k-1}^2 +
  z_{2k}^2}$. The windowing layer (third row of Table~\ref{tab:nfsc})
is a grouped convolution, meaning that each output filter only sees
the input filter with the same index. The decimation of $10$ms is
implemented in the stride of $160$ of this layer. Notice that to approximate the mel-filterbanks,
the square of the Hanning window is used and biases in both convolutional layers are set to zero.
We keep them to zero during training. We add log compression to the output of the grouped convolution after adding $1$ to its absolute value since we do not have positivity
constraints on the weights when learning. Contrarily to the MFSC, there
is no mean-variance normalization after the convolutions, but on the waveform. In the default implementation of the TD-filterbanks, we do not apply pre-emphasis. However, in our last experiment, we add a convolutional layer below the TD-filterbanks, with width $2$ and stride $1$, initialized with the pre-emphasis parameters, as another learnable component.

\vspace{-0.4cm}
\section{Experiments}
\label{sec:experiments}
\vspace{-0.4cm}
\subsection{Setting}
\vspace{-0.4cm}

We perform phone recognition
experiments on TIMIT \cite{timit} using the standard train/dev/test split. We train and evaluate our models with 39 phonemes. We experiment with three architectures. The first one consists of 5 layers of convolution of width 5 and 1000 feature maps, with ReLU activation functions, and a dropout \cite{dropout} of 0.5 on every layer but the input and output ones. The second model has the same architecture but a dropout of 0.7 is used. The third model has 8 layers of convolution, PReLU \cite{prelu} nonlinearities and a dropout of 0.7. All our models are trained end-to-end with the Autoseg criterion \cite{wav2letter}, using stochastic gradient descent. We compare all models using either the baseline MFSC as input or our learnable TD-filterbank front-end. We perform the same grid-search for both MFSC baselines and models trained on TD-filterbanks, using learning rates in $(0.0003,0.003)$ for the model and learning rates in  $(0.03, 0.003)$ for the Autoseg criterion, training every model for 2000 epochs. We use the standard dev set for early stopping and hyperparameter selection.
\vspace{-0.6cm}
\subsection{Different types of TD-filterbanks}
\vspace{-0.4cm}
Throughout our experiments, we tried four different settings for the TD-filterbank layers:
%\vspace{-0.2cm}
\begin{itemize}[leftmargin=*]

    \item Fixed: Initialize the layers to match MFSC and keep their parameters fixed when training the model
    \item Learn-all: Initialize the layers and let the filterbank and the averaging be learned jointly with the model
    \item Learn-filterbank: Start from the initialization and only learn the filterbank with the model, keeping the averaging fixed to a squared hanning window
    \item Randinit: Initialize the layers randomly and learn them with the network
\end{itemize}
Table \ref{tab:layer-modes} shows comparative performance of an identical architecture trained on the four types of TD-filterbanks. We can observe that training on fixed layers moderately worsens the performance, we hypothesize that this is due to the absence of mean-variance normalization on top of TD-filterbanks as is performed on MFSC. A striking observation is that a model trained on TD-filterbanks initialized randomly performs considerably worse than all other models. This shows the importance of the initialization. Finally, we observe better results when learning the filterbank only compared to learning the filterbank and the averaging but depending on the architecture it was not clear which one performs better. Moreover, when learning both complex filters and averaging, we observe that the learned averaging filters are almost identical to their initialization. %except for some noise that is added on the borders
Thus, in the following experiments, we choose to use the Learn-filterbank mode for the TD-filterbanks.
\vspace{-0.4cm}
\subsection{Results}
\vspace{-0.4cm}
We report PER on the standard dev and test sets of TIMIT. For each architecture, we can observe that the model trained on TD-filterbanks systematically outperforms the equivalent model trained on MFSC, even though we constrained our TD-filterbanks such that they are comparable to the MFSC and do not learn the low-pass filter. This shows that by only learning a new bank of 40 filters, we can outperform the MFSC for phone recognition. This gain in performance is obtained at a minimal cost in terms of number of parameters: even for the smallest architecture, the increase in number of parameters in switching from MFSC to TD-filterbanks is $0.31\%$. We also compare to baselines from the literature. One baseline trained on the waveform gets a PER of $29.1\%$ on the test set, which is in a range $8.8\%-11.1\%$ absolute above our models trained on the waveform. The Wavenet architecture, also trained on the waveform, yields a PER of $18.8$, which is higher than our best models despite using the phonetic alignment and an auxiliary prediction loss. Our best model on the waveform also outperforms a 2-dimensional CNN trained on MFSC and an LSTM trained on MFSC with derivatives. Finally, by adding a learnable pre-emphasis layer below the TD-filterbanks, we reach $18\%$ PER on the test set.
\vspace{-0.4cm}
\section{Analysis of learned filters}
\vspace{-0.4cm}

\begin{figure}
\centerline{\includegraphics[width=.5\textwidth]{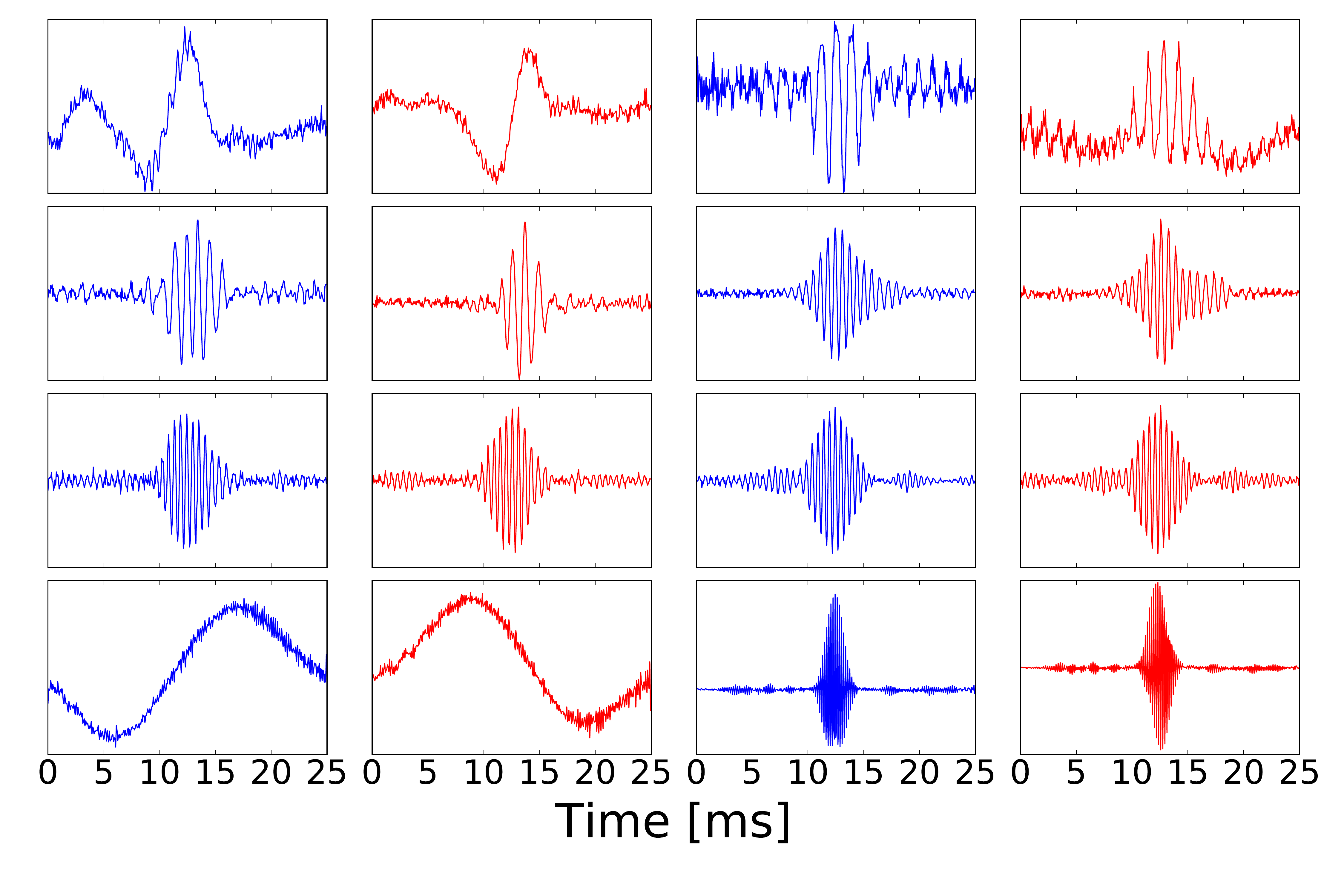}}
\vspace{-0.42cm}\caption{Examples of learned filters. Filters' real parts in blue; imaginary part in red. \label{fig:learnt_filters}}
\end{figure}

\begin{figure}
\centerline{\includegraphics[width=.5\textwidth]{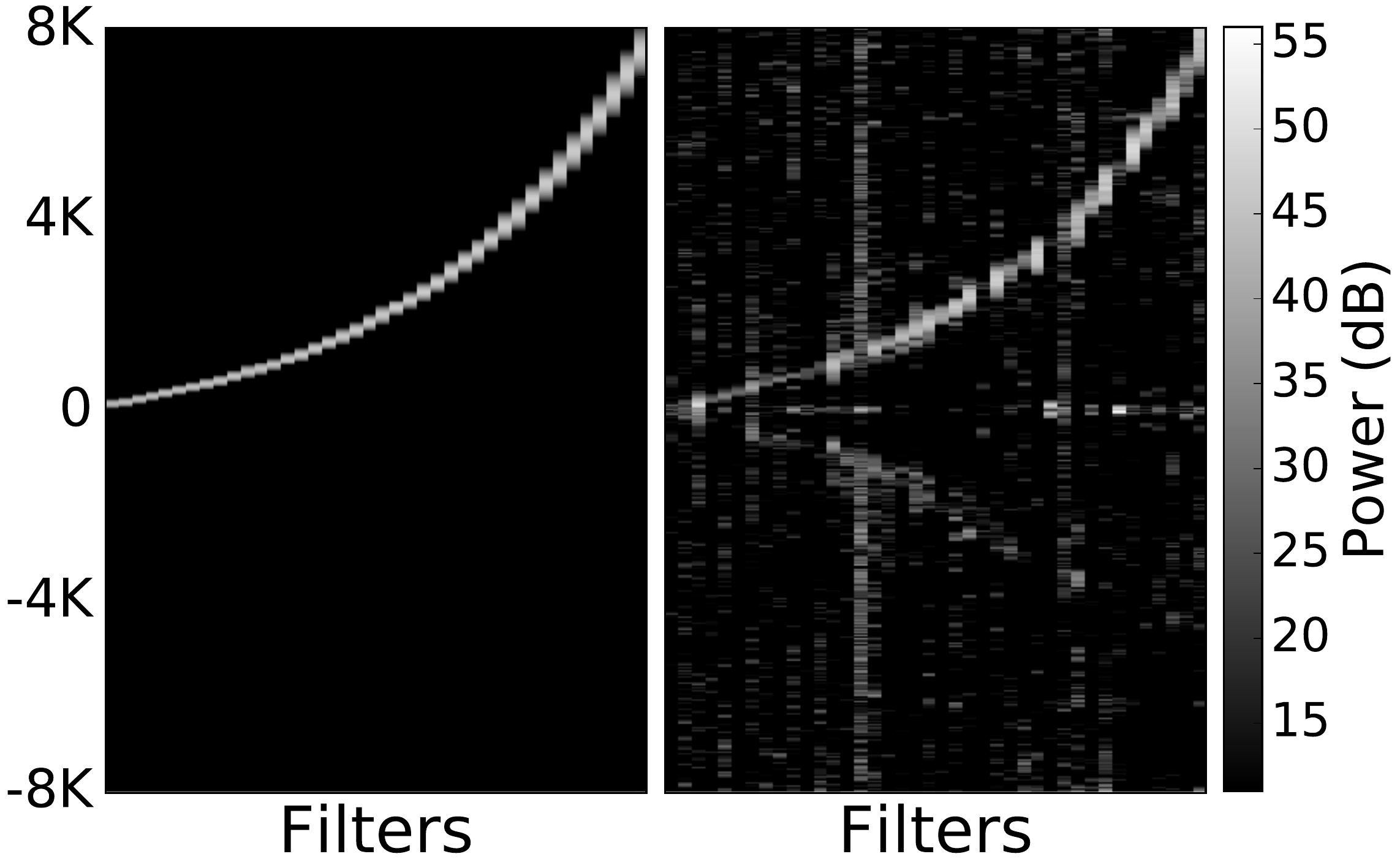}}
\vspace{-0.42cm}\caption{Heat-map of the magnitude of the frequency response  for initialization filters (left) and learned filters (right). \label{fig:filters_power_spectrum}}
\end{figure}

We analyze filters learned by the first layer of the CNN-8L-PReLU-do0.7 + TD-filterbanks model.
Examples of learned filters are shown in Figure \ref{fig:learnt_filters}. The magnitude of the frequency response for each of the 40 filters is plotted in Figure \ref{fig:filters_power_spectrum}. Overall, the filters tend to be well localized in time and frequency, and a number of filters became asymmetric during the learning process, with a sharp attack and slow decay of the impulse response. This is a type of asymmetry also found in human and animal auditory filters estimated from behavioral and physiological data \cite{smith2006efficient}. %
In Figure \ref{fig:filters_power_spectrum}, we further see that the initial mel-scale of frequency is mostly preserved, but that a lot of variability in the filter bandwidths is introduced.

A prominent question is whether the analyticity of the initial filterbank is preserved throughout the learning process even though nothing in our optimization method is biased towards keeping filters analytic. A positive answer would suggest that complex filters in their full generality are not necessary to obtain the increase in performance we observed. This would be especially interesting because, unlike arbitrary complex filters, analytic filters have a simple interpretation in terms of real-domain signal processing: taking the squared modulus of the convolution of a real signal with an analytic filter performs a sub-band Hilbert envelope extraction~\cite{flanagan1980parametric}.

A signal is analytic if and only if it has no energy in the negative frequencies. Accordingly, we see in Figure \ref{fig:filters_power_spectrum} that there is zero energy in this region for the initialization filterbank. After learning, a moderate amount of energy appears in the negative frequency region for certain filters. To quantify this, we computed for each filter the ratio $r_a$ between the energy in negative versus positive frequency components \footnote{Our model cannot identify if a given filter plays the role of the real or imaginary part in the associated complex filter. We chose the assignment yielding the smallest $r_a$.}. This ratio is 0 for a perfectly analytic filter and 1 for a purely real filter. We find an average $r_a$ for all learned filters of $.26$. Filters with significant energy in negative frequencies are mostly filters with an intermediate preferred frequency (between 1000Hz and 3000Hz) and their negative frequency spectrum appears to be essentially a down-scaled version of their positive frequency spectrum.

\vspace{-0.4cm}
\section{Conclusion}
\vspace{-0.4cm}
We proposed a lightweight architecture which, at initialization, approximates the computation of MFSC and can then be fine-tuned with an end-to-end phone recognition system. With a number of parameters comparable to standard MFSC, a TD-filterbank front-end is consistently better in our experiments. Learning all linear operations in the MFSC derivation, from pre-emphasis up-to averaging provides the best model.
In future work, we will perform large scale experiments with TD-filterbanks to test if a new state-of-the-art can be achieved by training from the waveform. 
\vspace{-0.4cm}
\section{Acknowledgements}
\vspace{-0.4cm}
Authors thank Mark Tygert for useful discussions, and Vitaliy Liptchinsky and Ronan Collobert for help on the implementation. This research was partially funded by the European Research Council (ERC-2011-AdG-295810 BOOTPHON), the Agence Nationale pour la Recherche (ANR-10-LABX-0087 IEC, ANR-10-IDEX-0001-02 PSL*).
% -------------------------------------------------------------------------
\begin{small}
\bibliographystyle{IEEEbib}
\bibliography{bib}
\end{small}
\end{document}